\documentclass[sigconf]{acmart}

\usepackage{svg}
\usepackage{multirow}
\usepackage{makecell}
\usepackage{bbding}
\AtBeginDocument{%
  \providecommand\BibTeX{{%
    \normalfont B\kern-0.5em{\scshape i\kern-0.25em b}\kern-0.8em\TeX}}}

\setcopyright{none}

\acmConference[Conference acronym 'XX]{Make sure to enter the correct
  conference title from your rights confirmation emai}{June 03--05,
  2018}{Woodstock, NY}
\begin{document}

\title{Multi-sentence Video Grounding for Long Video Generation}

% \affiliation{%
\author{Wei Feng}
\affiliation{%
Department of Computer Science and Technology, Tsinghua University
\city{Beijing}
\country{China}}
\email{fw22@mails.tsinghua.edu.cn}

\author{Xin Wang}
\authornote{Corresponding authors.}
\affiliation{%
Department of Computer Science and Technology, BNRist, Tsinghua University
  % \institution{Department of Computer Science and Technology, BNRist, Tsinghua University}
  \city{Beijing}
  \country{China}
  }
\email{xin_wang@tsinghua.edu.cn}

\author{Hong Chen}
\affiliation{%
Department of Computer Science and Technology, Tsinghua University
    \city{Beijing}
    \country{China}}
\email{h-chen20@mails.tsinghua.edu.cn}

\author{Zeyang Zhang}
\affiliation{%
Department of Computer Science and Technology, Tsinghua University
    \city{Beijing}
    \country{China}}
\email{zy-zhang20@mails.tsinghua.edu.cn}

\author{Wenwu Zhu}
\authornotemark[1]
\affiliation{%
Department of Computer Science and Technology, BNRist, Tsinghua University
  % \institution{Department of Computer Science and Technology, BNRist, Tsinghua University}
  \city{Beijing}
  \country{China}
  }
\email{wwzhu@tsinghua.edu.cn}
% \author{Anonymous Authors}
%% You do not have to enter your paper ID


\begin{abstract}
  Video generation has witnessed great success recently, but their application in generating long videos still remains challenging due to the difficulty in maintaining the temporal consistency of generated videos and the high memory cost during generation. To tackle the problems, in this paper, we propose a brave and new idea of Multi-sentence Video Grounding for Long Video Generation, connecting the massive video moment retrieval to the video generation task for the first time, providing a new paradigm for long video generation. The method of our work can be summarized as three steps: (i) We design sequential scene text prompts as the queries for video grounding, utilizing the massive video moment retrieval to search for video moment segments that meet the text requirements in the video database. 
(ii) Based on the source frames of retrieved video moment segments, we adopt video editing methods to create new video content while preserving the temporal consistency of the retrieved video. Since the editing can be conducted segment by segment, and even frame by frame, it largely reduces the memory cost.
% (ii) We propose a video generation method that combines structure preservation of source frames and consistency preservation among the generated video through, reconstructing all grounding video segments into video content with a unified subject or style.
(iii) We also attempt video morphing and personalized generation methods to improve the subject consistency of long video generation, providing ablation experimental results for the subtasks of long video generation.
Our approach seamlessly extends the development in image/video editing, video morphing and personalized generation, and video grounding to the long video generation, offering effective solutions for generating long videos at low memory cost.
\end{abstract}






\maketitle

\section{Introduction}
\label{sec:intro}

% With the rapid development of generative models, significant progress has been made in text-to-image and text-to-video generation.
% Diffusion Model(DM) has made significant progress in recent years, demonstrating the ability to generate images and videos.
% The main existing works of diffusion model focus on the video generation of short frame sequences, which typically include around 24 frames, limiting their application in the real world.

%Video Generation importance
Video generation has made significant progress in recent years, demonstrating the incredible ability to generate multimedia content.
The main existing works of video generation focus on developing different generative models, which can be divided into diffusion-based models~\cite{guo2023animatediff,blattmann2023stable,zhang2023i2vgen,wang2023modelscope} and non-diffusion-based models such as VQGAN~\cite{esser2021taming}. Yin et al. proposed NUWA-XL~\cite{yin2023nuwa} by applying Diffusion over Diffusion method for long video generation. In addition, some works strengthened temporal information by combining generative diffusion models or VQGAN models with the Transformers architectures to generate long videos of up to 1 minute~\cite{ge2022long,peebles2023scalable}. 
%Existing works, non-diffusion,  diffusion-based on diffusion models

%problem, frame, consistency, high memory cost

However, there are still many limitations in the generation of long videos.
The first issue is that the generated video content often overlooks some physical laws of real-world knowledge (such as chair running). In addition, the overall consistency of the generated video is lower than the real video due to unnatural transitions between frames. Last but not least, the longer the video is generated, the higher GPU memory cost would be required.

%propose a brave and new idea, grounding-based generation, which shares similar ideas with Retrieval-augmented generation. Specifically, generation, xxx, xxx. To ensure the consistency, we xxx. Experimental results, importance of this work, open new possibilities for long video generation.
To address these challenges, we propose a brave and new idea named grounding-based video generation, which applies the multi-sentence video grounding method for long video generation. This idea shares similar spirits with the retrieval-augmented generation{~\cite{asai2023self,yoran2023making,gao2023retrieval}} in large language models. To begin with, based on the video grounding technique such as massive video moment retrieval, we can obtain several moments of different videos from our video database that match target text queries, to provide video generation tasks with guided source video segments that follow physical rules and remain highly consistent. The retrieved video segments will provide motion information for the final generated video. Subsequently, based on the retrieved video segments, we adopt the video editing method to create new content in the video segments, such as changing the subject or changing the background. Additionally, we combine the edited video segments with a unified subject or style through video editing, and achieve long video generation while ensuring overall consistency and adherence to the physical laws of the generated video content. Meanwhile, considering that video editing can be conducted segment by segment and even frame by frame, our work maintains a relatively low level of GPU memory cost, making it possible for the public to generate long videos. Extensive experimental results show that our proposed method can be used to generate long videos with better consistency. In future works, with a larger video corpus and more advanced video grounding methods, our proposed method can work as a powerful long video generation tool.

To summarize, we make the following contributions:

\begin{itemize}
    \item To the best of our knowledge, this is the first work to study the feasibility of leveraging the multi-sentence video grounding for long video generation, which we believe will inspire a lot of future work.
    \item We propose the Multi-sentence Video Grounding-based Long Video Generation framework, consisting of i) a massive video moment retrieval model capable of locating suitable video segments for the text prompts, ii) a video editor that creates new content for the video segments while preserving temporal consistency and iii) a video personalization and morphing scheduler that enables customized video generation and smooth transition between generated videos.
    \item We conduct experiments on various video editing and video personalization methods, demonstrating the feasibility of retrieval augmentation to improve the continuity and diversity of generated long videos through the video grounding method.
    % \ch{demonstrating}.
    % \item \ch{We provide xxx analysis}
    \item We conduct ablation analysis under different video editing methods and the application of video morphing and personalization, providing importing references for improving the performance of long video generation.
\end{itemize}

\section{Related work}
\label{relate}

\subsection{Video Grounding}
Video grounding aims to locate the starting and ending times of a given segment target from a video\cite{chen2018temporally,feng2023multimedia,nan2021interventional}, which is a popular computer vision task and has drawn much attention over the past few years~\cite{yuan2021closer,feng2023llm4vg}. Early-stage video grounding task mainly focused on searching for target segments from a single video, which limited its ability to obtain information from the entire video pools. Therefore, tasks such as video corpus moment retrieval (VCMR)~\cite{lei2020tvr,ma2022interactive,jung2022modal} have emerged in the field of computer vision, which focuses on finding one correct positive video segment associated with the VG query from the video pools.

Furthermore, the Massive Videos Moment Retrieval (MVMR)~\cite{yang2023mvmr} task was proposed, which assumes that there could be several positive video segments in the video pools that match a certain target query. This task requires the model to distinguish positive from massive negative videos. To address this challenge, Yang et al. introduce the Reliable Mutual Matching Network (RMMN)~\cite{yang2023mvmr}, which mutually matches a query with representations of positive video moments while distinct from negative ones.  

\subsection{Long Video Generation}

Existing work has demonstrated impressive abilities in generating high-quality images and short videos~\cite{podell2023sdxl,balaji2022ediff,meng2021sdedit,rombach2022high,guo2023animatediff,he2023animate}. By introducing the transformer architecture to enhance temporal understanding and reasoning ability, some work has been able to generate long videos.
~\citet{villegas2022phenaki} present Phenaki with C-ViViT as encoder and MaskGit as the backbone, which is able to generate variable-length videos conditioned on a sequence of prompts in the open domain. With the rapid development of Diffusion Models (DM)~\cite{ho2020denoising,song2020score} such as stable diffusion~\cite{rombach2022high}, ~\citet{peebles2023scalable} proposed Diffusion Transformers (DiTs), a simple
transformer-based backbone for diffusion models that outperform prior U-Net models. Given the promising scaling results of DiTs, OpenAI proposes Sora, presenting powerful abilities to generate long videos and simulate the physical world.

These works, however, all have limitations when addressing long video generation tasks. On the one hand, a common problem in generating videos is that some content may violate physical knowledge or have poor overall video consistency.
The reasons causing this problem, include inadequate prompt understanding of input and the limitations of the generation algorithm, leading to insufficient modeling of physical laws. On the other hand, for generative models that consider the overall temporal relationship of video generation and apply methods such as the Transformer architecture, to generate a single long video at once, the problem they face is that the longer the video is generated, the more computational resources are needed.

\subsection{Video Editing}

Benefiting from the rapid development of diffusion models in image and video generation, many zero-shot video editing methods have been proposed~\cite{cong2023flatten}, which apply the pre-trained image diffusion model to transform an input source video into a new video. The critical problem of video editing is to maintain the visual motion and temporal consistency between the generated video and the source video. 

To address these problems, some works introduce additional spatial conditioning controls or internal features to keep motion consistency between the generated video images and the source video images. LOVECon~\cite{liao2023lovecon} applies ControlNet~\cite{zhang2023adding} using conditions control such as edges, depth, segmentation, and human pose for text-driven training-free long video editing. Text2LIVE~\cite{bar2022text2live} generates an edit layer with color and opacity features to constrain the generation process and demonstrate edits on videos across a variety of objects and scenes.
In addition, other methods are proposed to improve the temporal coherency in generation by considering the relationship between source video frames. Tune-A-Video~\cite{wu2023tune} proposes a One-Shot Video Tuning on the source video and appends temporal self-attention layers for consistent T2V generation. Pix2Video~\cite{ceylan2023pix2video} applies feature injection and latent update methods, using the intermediate diffusion features of a reference frame to update the features of future frames.
VidToMe~\cite{li2023vidtome} unifies and compresses internal features of diffusion by merging tokens across video frames, balancing the performance of short-term video continuity and long-term consistency of video editing.

\section{method}
\begin{figure*}[h]
    \centering
    \includegraphics[width=\linewidth]{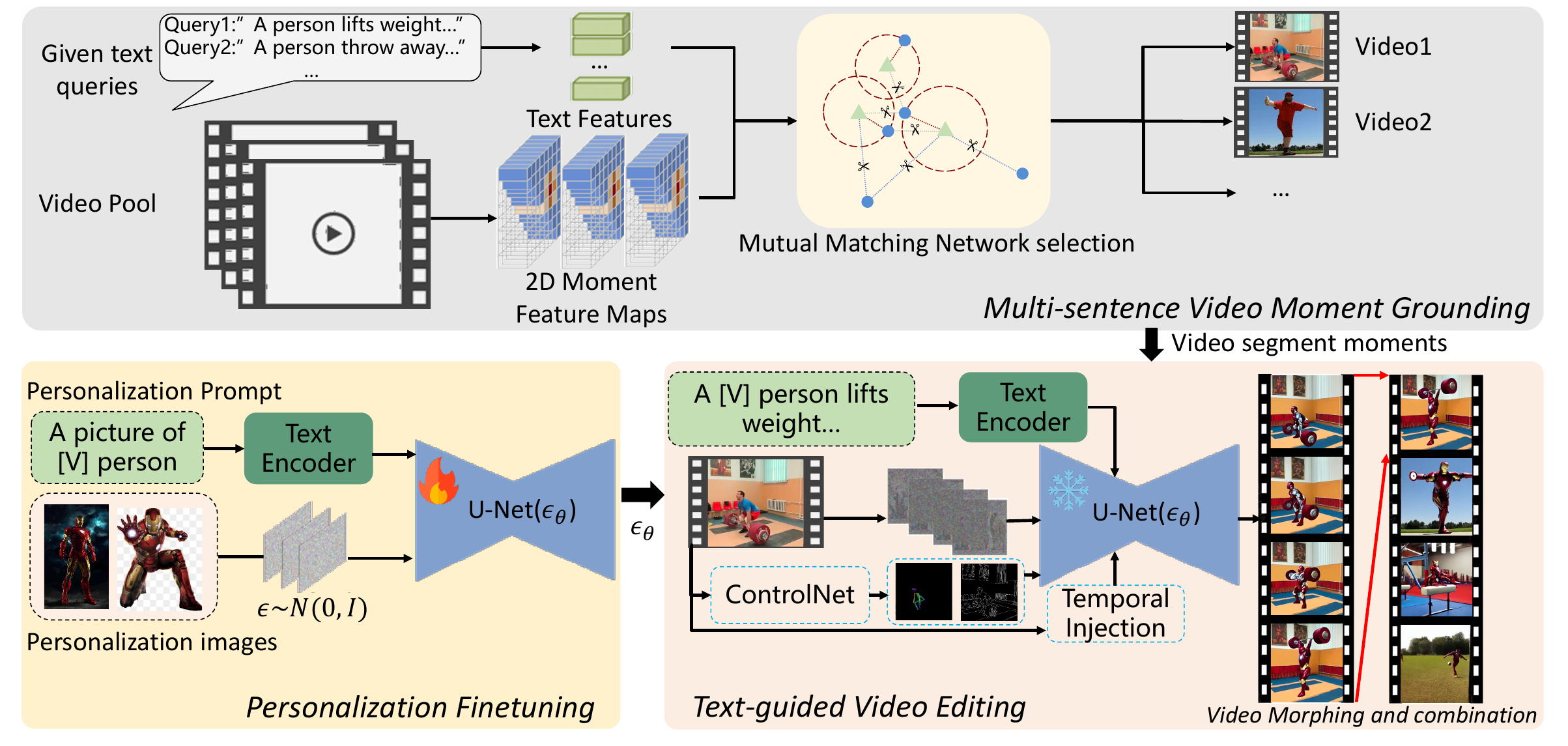}
    \vspace{0.2cm}
    \caption{Framework of Multi-sentence Video Grounding for Long Video Generation. In the stage of Multi-sentence Video Moment Grounding, we input a sentence of queries $(q_1,q_2,...,q_n)$ and obtain their corresponding video segments $V_1, V_2,..., V_n=Grounding(q_1,q_2,...,q_n)$. In the stage of Text-guided Video Editing, each received video segment would go through video editing and form the generated video $V'=Editing(V,q')$ with a unified subject or scenario. The obtained edited videos $V'_1, V'_2,..., V'_n$ would be smoothly combined into a long video using the Video Morphing method. The Personalization Finetuning is optional to replace the diffusion model for generating videos with customized subjects.}
    \label{fig:framework}
    \vspace{0.3cm}
\end{figure*}

Long video generation task requires generating a consistent and diverse video of at least one minute conditioned on a story or a sequence of prompts $(p_1,p_2,...,p_n)$.
% Given a sequence with several target queries $(q_1,q_2,...,q_n)$, each query in the format of \textit{`A person does something...'}, long video generation task requires to utilize their modified queries $(\hat{q_1},\hat{q_2},...,\hat{q_n})$ in the format of \textit{`\textbf{Customized subject} does something...'} to generate a sequence of consistent videos, where \textit{`\textbf{Customized subject}} could be any personalized subject, and combine the generated videos into one video with smooth transition.
As shown in Figure~\ref{fig:framework},
the method of Multi-sentence Video Grounding for Long Video Generation can be decomposed into the following steps. 

Given a sequence with several target queries $(q_1,q_2,...,q_n)$, each query in the format of \textit{`A person does something...'},
% The first step is to design sequential original text prompts as the queries $(q_1,q_2,...,q_n)$ in the format of \textit{`A person does something...'} for video grounding,
our first step is to input them into the Massive Video Moment retrieval model to search for video time segments that meet the requirements in the video database and filter a sequence of videos $V_1, V_2,..., V_n=Grounding(q_1,q_2,...,q_n)$. Secondly, we will use their modified queries $({q'_1},{q'_2},...,{q'_n})$ in the format of \textit{`\textbf{Customized subject} does something...\textit{\textbf{in a customized scenario}}'} and the video editing method to edit each grounding video segment into video content $V'=Editing(V,q')$ with a unified subject we want to customize and change the background as we expected, forming several story segments $V'_1, V'_2,..., V'_n$ of a continuous long video.
In addition, we also attempt video morphing approaches for the combination of generated video segments, and personalized generation methods to improve the subject consistency of long video generation.

\subsection{Multi-sentence Video Moment Grounding}
Given a sequence with several target queries $(q_1,q_2,...,q_n)$ in the format of \textit{`A person does something...'}, we first tokenize each query $q$ by DistillBERT and then perform global average pooling over all the tokens to transform them into text feature$f^q\in R^d$. DistillBERT~\cite{sanh2019distilbert} is a lightweight text query encoder showing comparable performance to BERT~\cite{kenton2019bert} but with a smaller size and faster computational speed. We hope to find video clips from a video pool that match the text feature through the video grounding method.

For each video in the database, we divided it into $N$ video clips and used a pre-trained visual model C3D~\cite{tran2015learning} for feature extraction. Utilizing these features through FC layer dimensionality reduction and max pooling, we constructed a 2D temporary moment feature map $F\in R^{N\times N\times d}$.

After transforming each query and video moment using encoders to a joint visual-text space, we apply Mutual Matching Network(MMN)~\cite{wang2022negative} trained by the Cross-directional Hard and Reliable Negative Contrastive Learning (CroCs) method~\cite{yang2023mvmr} to solve the Multi-sentence Video moment grounding task. From the joint visual-text space, the matching score between the query and the video moment representations could be computed through cosine similarity between them. 

\begin{equation}
\begin{split}
f^q_{mm}&=W_{mm}f^q+b_{mm},\\
    \forall {f^v}&\in F, s^{mm} = {f^v}^T_{mm}f^q_{mm},
\end{split}
\end{equation}
where $W_{mm}$ and $b_{mm}$ are learnable parameters and we enforce the embeddings $||f^v_{mm}||=||f^q_{mm}||=1$ through a $l_2$-normalization layer.

The video grounding segments from the few videos with the highest match $s_{mm}$ with each query would be selected for subsequent editing.

\subsection{Text guided Video Editing}
\label{sec:reconstruct}
After receiving video segments $V_1, V_2,..., V_n=Grounding(q_1,q_2,...,q_n)$ from the Multi-sentence Video Moment Grounding approach.
For each video clip $V$ with $n$ frames$(v^1,v^2,...,v^n)$ corresponding to a text query $q$, each frame would be encoded into a low-dimensional latent representation $z^i_0=E(v^i)$, and go through DDIM Inversion, converting them back into noise latent in $T$ reverse steps:
\begin{equation}
\begin{split}
    z^i_{t+1}&=\sqrt{\alpha_{t+1}}\frac{z^i_t-\sqrt{1-\alpha_t}\epsilon_\theta(z^i_t,t,c_q)}{\sqrt{\alpha_t}}+\sqrt{1-\alpha_{t+1}}\epsilon_\theta(z^i_t,t,c_q),\\
    t &= 0,...,T-1,
\end{split}
\end{equation}
where $\epsilon_\theta$ represents an image-to-image translation U-Net of the diffusion models we implement, $\alpha_t $ represents the noise variance based on the decreasing schedule and $c_q$ is the text embedding encoded from the text query $q$.

Using the noise latent as a starting point for video editing, we modify each original video query $q$ to a new query ${q'}$, making different text queries have the same character subject or visual style.
The basic video frame editing approach uses the following DDIM Sampling methods to directly edit each video frame:
    
\begin{equation}
\begin{split}
    z^i_{t-1}&=\sqrt{\alpha_{t-1}}\frac{z^i_t-\sqrt{1-\alpha_t}\epsilon_\theta(z^i_t,t,c_{q'})}{\sqrt{\alpha_t}}+\sqrt{1-\alpha_{t-1}}\epsilon_\theta(z^i_t,t,c_{q'}),\\
    t &= 0,...,T-1,
\end{split}
\end{equation}
while this method, however, would be unable to consider the temporal consistency between the video frames. The quality of edited images is mainly limited by the diffusion model and the detail level of the prompt query.

Therefore, we introduce additional methods that improve video editing and generation from two aspects. On the one hand, we use ControlNet which adds conditional control such as edges, depth, segmentation, and human pose as reference information to the Diffusion Model for better generation guidance. On the other hand, we modify the intermediate latent through video editing approaches such as pre-frame latent injection, cross-window attention, and global token merging. Given these methods mentioned above, we can update each latent $z_t^i$ with richer reference information from pre-step latent and source video images.

In the end, each final latent would be mapped back to an image frame through a decoder $f^i = D(z^i_0)$, forming the generated video $V'=Editing(V,q')$ with higher temporal consistency that more physically makes sense.

\subsection{Video Morphing and Personalization}
\subsubsection{Video Morphing}
Although we have obtained edited videos of several different scenes $V'_1, V'_2,..., V'_n$ with consistent character subjects or styles through video grounding and generative methods, there are still significant differences between the videos based on different text queries, and they cannot be smoothly combined into a long video since the end frame of the previous video could not be not coherent with the starting frame of the next video.

Therefore, we adopted the video morphing method to concatenate the start and end segments of all edited videos. In each transition task, we obtain the VAE encoded latent of the preceding video's last frame $v^i$ and the following video's first frame $v^j$, represented as $z^i_0$ and $z^j_0$, along with the modified queries ${q'_i}$ and ${q'_j}$ of the two videos. Inspired by the approach of DiffMorpher~\cite{zhang2023diffmorpher}, we use two sets of latent-query pairs to relatively fine-tune the diffusion model and train two LoRAs~\cite{hu2021lora} $\Delta\theta_i$ and $\Delta\theta_j$ on the SD UNet $\epsilon_\theta$ according to the following learning objective:
\begin{equation}
\begin{split}
    L(\Delta\theta)=E_{\epsilon,t}[||\epsilon-\epsilon_{\theta+\Delta\theta}(\sqrt{\alpha_t}z_0+\sqrt{1-\alpha_t}\epsilon,t,c_q)||^2],
\end{split}
\end{equation}
where $\epsilon\sim N(0,I)$ is the random sampled Gaussian noise.

After fine-tuning, $\Delta\theta_i$ and $\Delta\theta_j$ are fixed and fused into a linear interpolation for the semantics of the input images:
\begin{equation}
    \Delta\theta_\alpha=(1-\alpha)\Delta\theta_i+\alpha\Delta\theta_j, 
\end{equation}
where $\alpha=\frac{k}{n},k=1,2,...,n-1$, representing generation process of the k-th transition image latent from ${q'_i}$ to ${q'_j}$, we take $n$ as 15. 

Using LoRA fine-tuned by the images, we utilize LoRA-integrated UNet $\epsilon_{\theta+\Delta\theta_k}(k=i,j)$  to relatively inverse image latent $z^i_0$ and $z^j_0$ to $z^i_T$ and $z^j_T$, and obtain the intermediate latent noise $z^\alpha_T$ through spherical linear interpolation~\cite{shoemake1985animating}:

\begin{equation}
\begin{split}
    z^\alpha_T &= \frac{sin((1-\alpha)\phi)}{sin\phi}z^i_T+\frac{sin(\alpha\phi)}{sin\phi}z^j_T,\\
    \phi &= arccos(\frac{z^i_T\cdot z^j_T}{||z^i_T|| ||z^j_T||}).
\end{split}
\end{equation}

When generating the intermediate image latent $z^\alpha_0$, we use the UNet with interpolated LoRA $\epsilon_{\theta+\Delta\theta_\alpha}$ during the DDIM Sampling steps, while the latent condition $c_\alpha = (1-\alpha)c_{{q'_i}}+\alpha c_{{q'_j}}$ is applied through the linear interpolation. By generating $z^\alpha_0, \alpha=\frac{k}{n},k=1,2,...,n-1$ and $v^\alpha = D(z^\alpha_0)$, we formed a transition video segment $V^{ij}=(v^\frac{1}{n},v^\frac{2}{n},...,v^\frac{n-1}{n})$ from $v^i$ to $v^j$.

\subsubsection{Video Personalization}
Video personalization is designed to generate characters from the real or virtual world that were not used for training the diffusion model.
In our approach, we fine-tune the text-to-image diffusion model using 3-5 images of a specialized character paired with the text prompt containing a rare token identifier~\cite{ruiz2023dreambooth} and the name of the character's class(e.g., "A [V] dog" or "A [sks] man").

After fine-tuning, we could replace the diffusion model $\epsilon_\theta$ used in sec~\ref{sec:reconstruct} and video morphing. This step is optional to generate customized subjects for the final generated videos.

\section{Experiments}
In this section, we will present the results of our method on long video generation tasks based on multi-sentence video grounding and generative methods.

\begin{figure*}[!htpb]
    \centering
    \includegraphics[width=\linewidth]{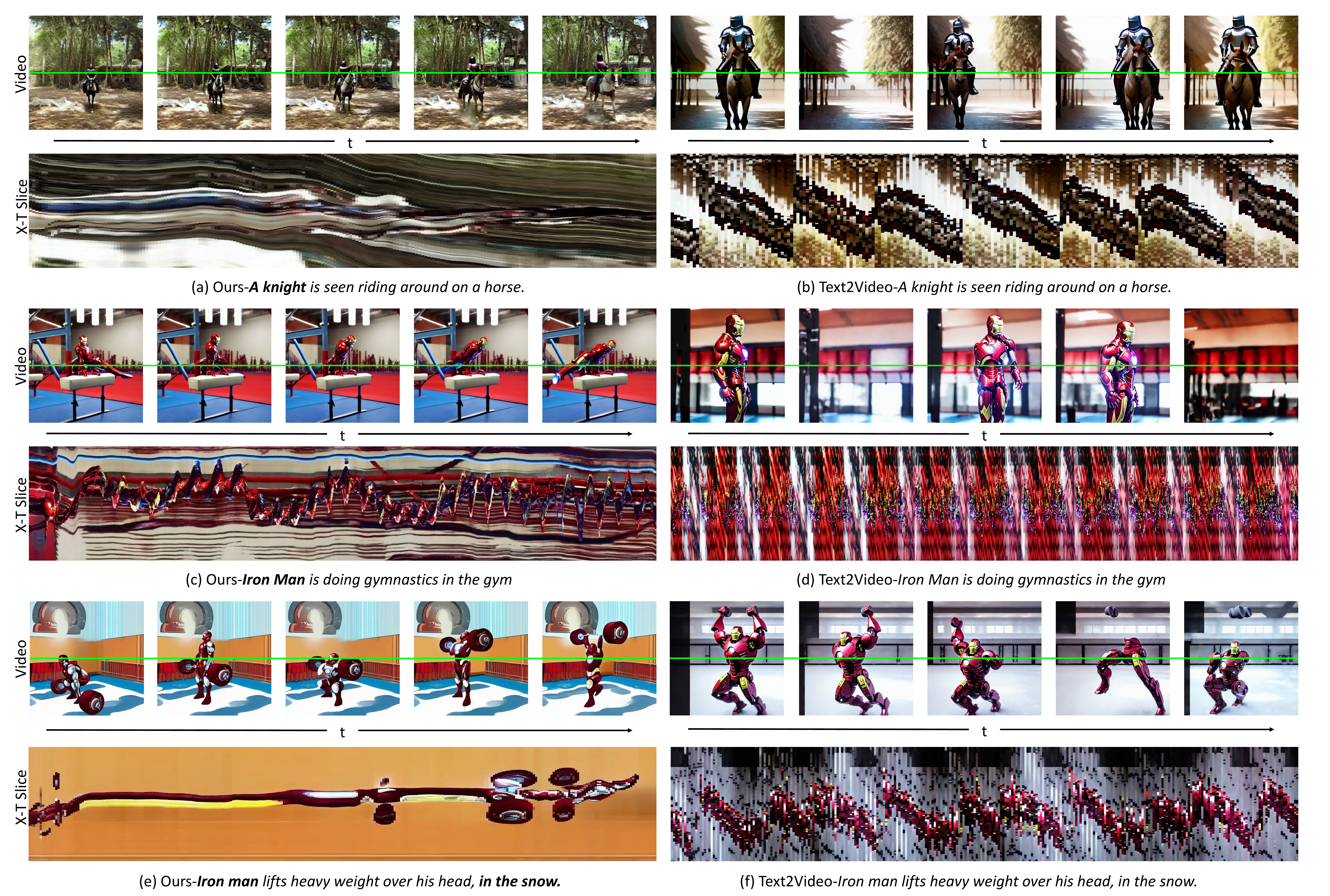}
    \vspace{0.3cm}
    \caption{Qualitative example results. (a),(c) and (e) are example videos generated through our method, including the customized subject or scenario in the text queries represented by the bold characters. While (b), (d), and (f) are videos generated through the baseline model.}
    \label{fig:example}
\end{figure*}

\subsection{Setups}
% \noindent\textbf{Datasets. }
% \paragraph{Datasets.}
\subsubsection{Datasets.} We conduct massive video moment retrieval on the Charades~\cite{gao2017tall}, ActivityNet~\cite{caba2015activitynet}, and TACoS~\cite{regneri2013grounding} datasets and mainly choose \textbf{Activitynet} as our video database for the next stage of video editing since it covers a wide range of complex human activities that are of interest to people in their daily living. We generate about 100 edited videos for evaluation, each containing frames ranging from 150 to 1000 (depending on the length of the video clip captured by video grounding) with a resolution of $512\times512$.

\subsubsection{Models. } We apply Stable Diffusion model with video editing approaches including \textbf{Pix2Video}, \textbf{LOVECon} and \textbf{VidToMe} for the video editing. These video editing methods apply different methods to enhance the temporal consistency such as Pix2Video uses self-attention feature injection to propagate the changes to the future frames, LOVECon proposes a cross-window attention mechanism to ensure the global style, and VidToMe enhances both short-term and long-term temporal consistency by merging self-attention tokens across frames. In addition, we also conduct video editing with \textbf{ControlNet} of various conditioning controls to improve motion consistency.

\subsubsection{Baseline. } Considering that our work is built on the Stable Diffusion model for video generation, we compare our results with the Stable Diffusion based video generation method \textbf{Text2Video-Zero}~\cite{khachatryan2023text2video} that encodes motion dynamic in the latent codes and reprograms cross-frame attention of frames, allowing zero-shot text-to-video generation. 
% Our approach has the same low memory cost as it, which focusing on DDIM conversion and DDIM sampling

\begin{table*}[h]
% \small
    \centering
    \caption{Results with different editing methods of our approach.} 
    \begin{tabular}{cccccccc}
    \toprule
        Editing Method&ControlNet&subject consistency&imaging quality&temporal style&temporal flickering\\
    \midrule
    Pix2Video&None&73.07\%&61.69\%&10.16\%&98.41\%\\
    % \multirow{3}{*}{LoveCON}
   LoveCon&Canny&70.99\%&53.65\%&8.39\%&98.49\%\\
       LoveCon&Depth&71.60\%&54.61\%&9.21\%&98.18\%\\
   LoveCon&Hed&70.83\%&54.94\%&9.22\%&98.48\%\\
   
   VidToMe&None&\textbf{76.79\%}&\textbf{69.50\%}&10.59\%&\textbf{98.67\%}\\
   VidToMe&Depth&72.90\%&63.33\%&10.58\%&98.43\%\\
   VidToMe&Canny&73.43\%&66.17\%&\textbf{10.85\%}&97.59\%\\
   VidToMe&Softedge&72.19\%&63.66\%&10.08\%&98.26\%\\
    \bottomrule
    \end{tabular}
    \vspace{0.5cm}
    
    \label{tab:total}
    
\end{table*}

\subsubsection{Evaluation Metrics. } Inspired by VBench~\cite{huang2023vbench}, we conduct measurements on the generated long video content as follows: (i) The first aspect we evaluate is \textbf{subject consistency}, representing whether the appearance of the main subject remains consistent across the long video, which is calculated by the DINO~\cite{caron2021emerging} score based on the frames' feature similarity. (ii) Generating video content often results in distortion in frames, so we use the MUSIQ~\cite{ke2021musiq} predictor trained on the SPAQ~\cite{fang2020perceptual} dataset to evaluate the \textbf{image quality} of the video. (iii) \textbf{Temporal style} represents the continuity of camera motion and, like subject consistency, is a part of the video's overall consistency. We evaluate it by calculating the similarity between the video feature and the temporal style description feature through ViCLIP~\cite{wang2023internvid}. (iv) For \textit{local and high-frequency details}, \textbf{temporal flickering} requires the video to possess imperfect temporal consistency instead of being static images. We use the mean absolute difference across frames to evaluate it.

\begin{figure*}[h]
    \centering
    \includegraphics[width=\linewidth]{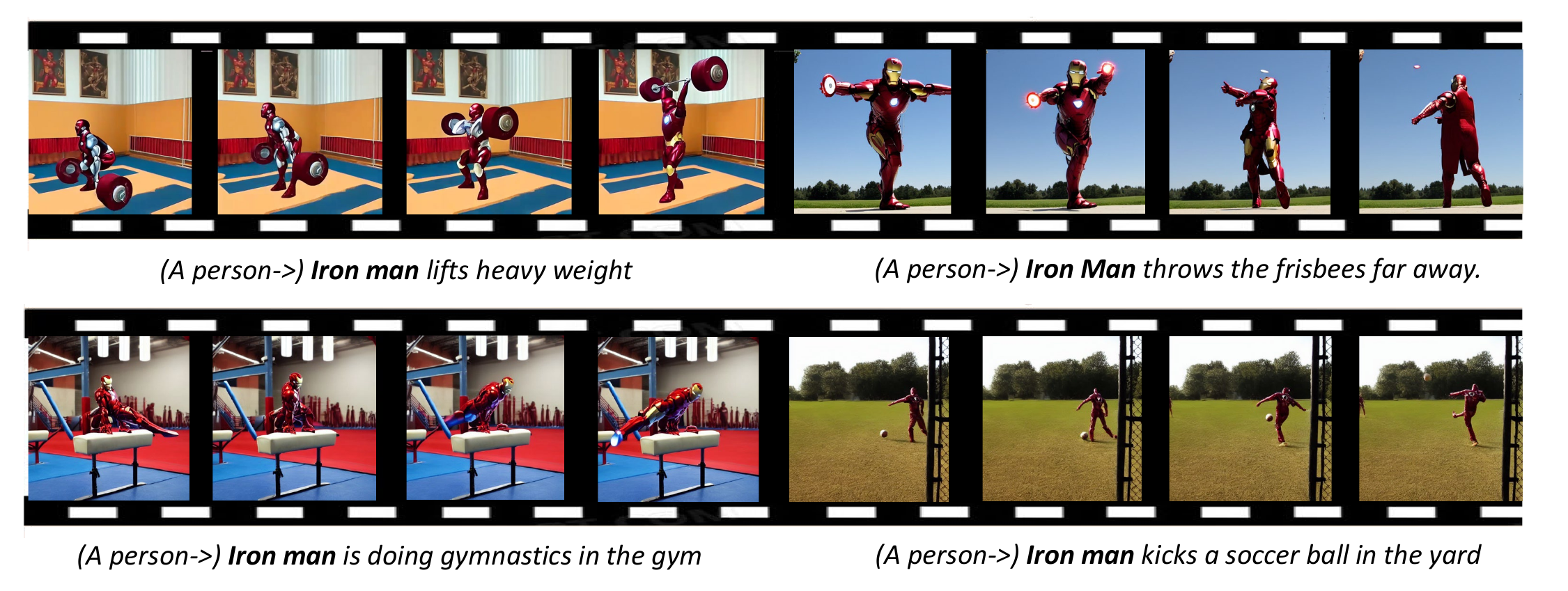}
    \vspace{0.1cm}
    \caption{Example combined video using multi-sentence video grounding for long video generation. The non-bold texts represent queries for video grounding, while bold text represents a portion of the content in the query being replaced in video editing to generate a customized subject.}
    \vspace{0.2cm}
    \label{fig:positive}
\end{figure*}

\begin{figure}[!htpb]
    \centering
    \includegraphics[width=1.03\linewidth]{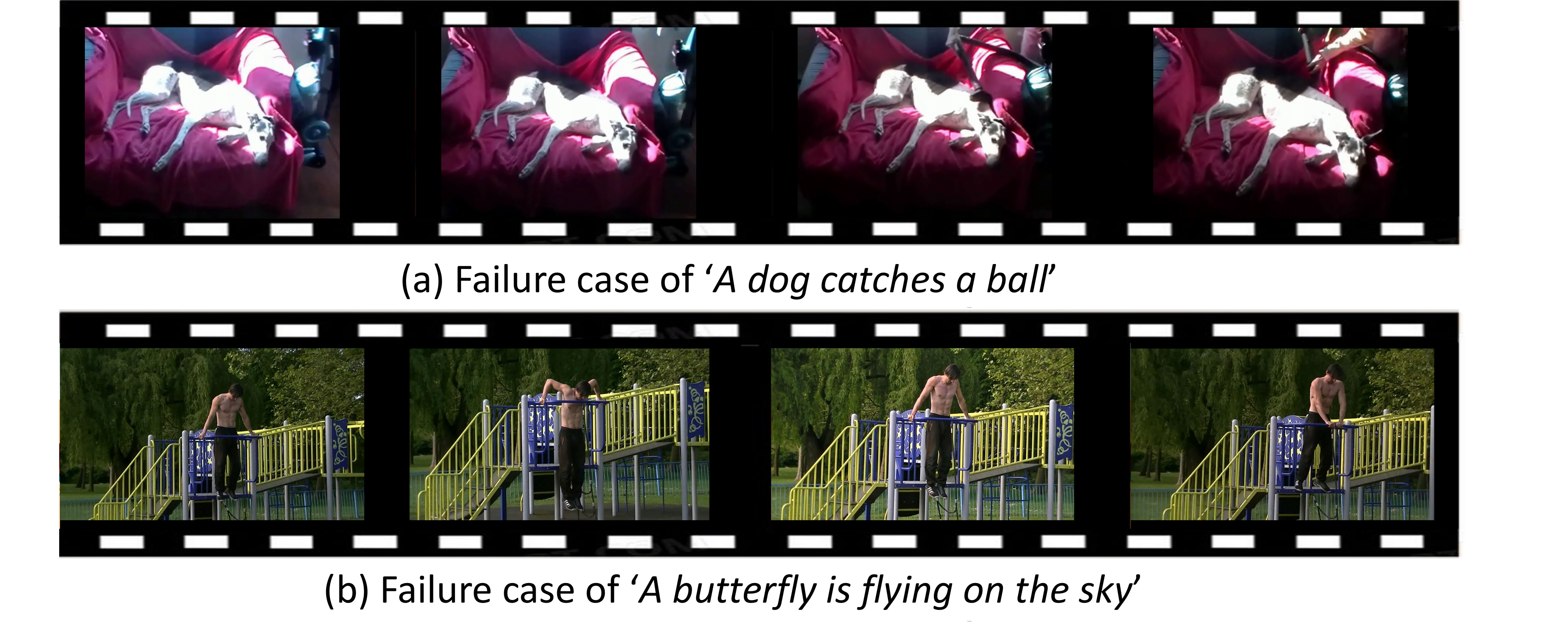}
    
    \caption{Failure video grounding examples. The video grounding model fails to retrieve the correct video segments that exist in the video dataset.}
    \label{fig:failure1}
\end{figure}
\begin{figure}[!htpb]
    \centering
    \includegraphics[width=0.96\linewidth]{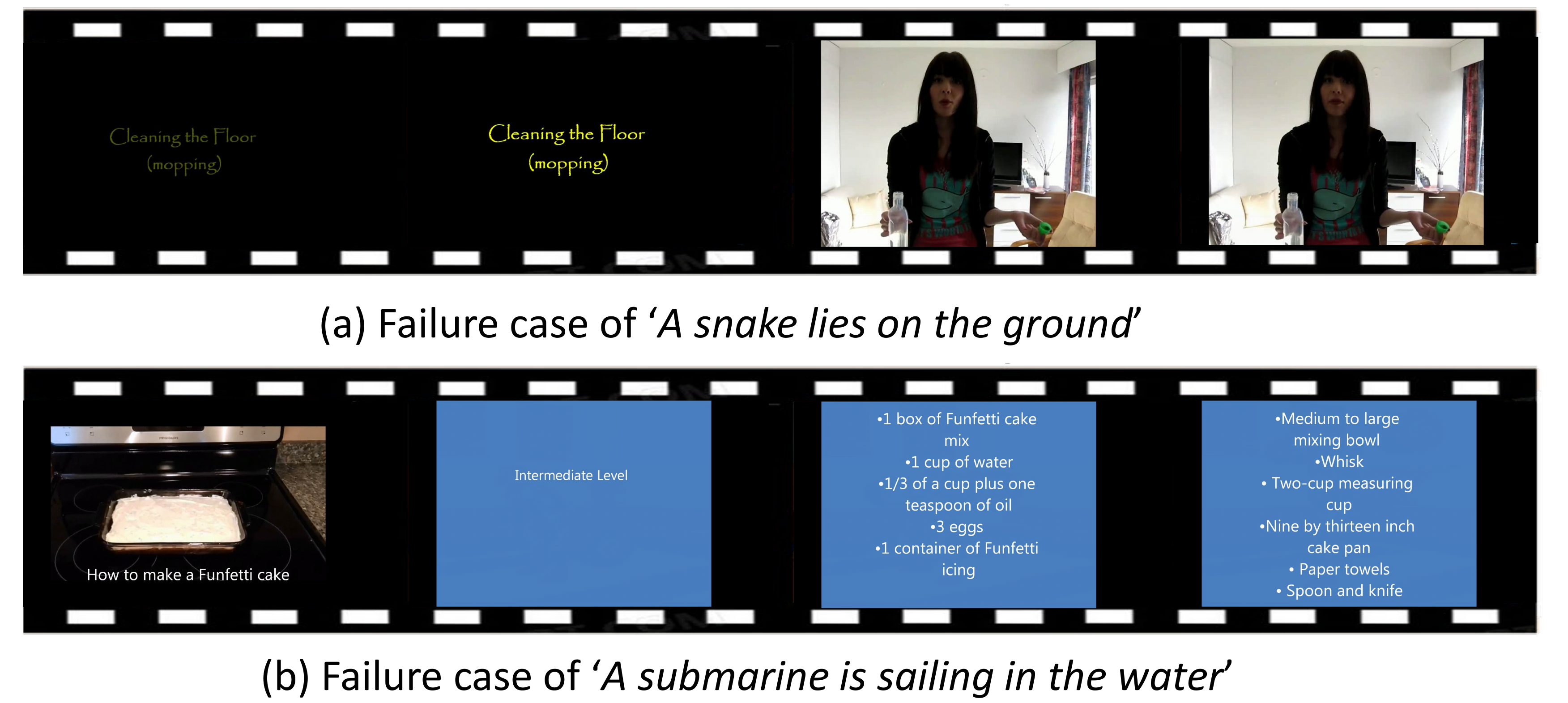}
    \caption{Failure video grounding examples. The video grounding dataset lacks video segments of some specific subjects such as \textit{snake} or \textit{submarine}.}
    \label{fig:failure2}
\end{figure}

\begin{table}[htbp]
\small
    \centering
    \caption{Comparison between our method with baseline.} 
    \begin{tabular}{cccccccc}
    \toprule
        \multirow{2}{*}{Model}&\multirow{2}{*}{\makecell{subject \\consistency}}&\multirow{2}{*}{\makecell{imaging \\quality}}&\multirow{2}{*}{\makecell{temporal\\ style}}&\multirow{2}{*}{\makecell{temporal\\flickering}}\\
        \\
    \midrule
    Text2Video-Zero&68.76\%& 62.18\%& 10.46\%& 80.10\%\\
    % \multirow{2}{*}{\makecell{Ours\\(with Pix2Video)}}&\multirow{2}{*}{73.07\%}&\multirow{2}{*}{61.69\%}&\multirow{2}{*}{10.16\%}&\multirow{2}{*}{98.41\%}\\\\
    Ours with Pix2Video&73.07\%&61.69\%&10.16\%&98.41\%\\
    % \multirow{3}{*}{LoveCON}
    Ours with LoveCon&70.83\%&54.94\%&9.22\%&98.48\%\\
   
   Ours with VidToMe&\textbf{76.79\%}&\textbf{69.50\%}&\textbf{10.59\%}&\textbf{98.67\%}\\
   
    \bottomrule
    \end{tabular}
    \vspace{0.5cm}
    
    \label{tab:main}
    
\end{table}

\begin{table}[!htbp]
\footnotesize
    \centering
    \caption{Ablation study for the application of Video Morphing and Personalization.} 
    \begin{tabular}{cccccccc}
    \toprule
        \multirow{2}{*}{\makecell{Video \\Morphing}}&\multirow{2}{*}{Personalization}&\multirow{2}{*}{\makecell{subject \\consistency}}&\multirow{2}{*}{\makecell{imaging \\quality}}&\multirow{2}{*}{\makecell{temporal\\ style}}&\multirow{2}{*}{\makecell{temporal\\flickering}}\\
        \\
    \midrule
    \multicolumn{2}{c}{Baseline}&68.76\%& 62.18\%& 10.46\%& 80.10\%\\
    \midrule
    \XSolidBrush&\XSolidBrush&72.90\%&63.33\%&10.58\%&98.43\%\\
    % \multirow{2}{*}{\makecell{Ours\\(with Pix2Video)}}&\multirow{2}{*}{73.07\%}&\multirow{2}{*}{61.69\%}&\multirow{2}{*}{10.16\%}&\multirow{2}{*}{98.41\%}\\\\
    \Checkmark&\XSolidBrush&72.81\%&64.61\%&10.74\%&98.33\%\\
    % \multirow{3}{*}{LoveCON}
    \XSolidBrush&\Checkmark&72.47\%&63.36\%&10.03\%&98.45\%\\
   
   \Checkmark&\Checkmark&72.43\%&64.60\%&9.46\%&98.35\%\\
   
    \bottomrule
    \end{tabular}
    \vspace{0.5cm}
    
    \label{tab:ablation}
    
\end{table}

\subsection{Main results}
% \noindent\textbf{Quantitative results. } 
Our main quantitative results of video grounding for long video generation are shown in Table~\ref{tab:main}. From the results, we can observe that:
% \begin{itemize}
%     \item Our video grounding-based long video generation method achieves higher scores stably in subject consistency and temporal flickering, demonstrating the feasibility of retrieval augmentation to improve the continuity and diversity of generated long videos through video grounding methods.
%     \item In terms of image quality and temporary style performance, the best performance our method achieves is higher than the baseline method, indicating that it can improve the generated videos' image quality. However, these improvements are not significant, so it is worth further exploring ways to improve performance by introducing other image generation models or other temporal consistency video editing methods.
% \end{itemize}
(i) Our video grounding-based long video generation method achieves higher scores stably in subject consistency and temporal flickering, demonstrating the feasibility of retrieval augmentation to improve the continuity and diversity of generated long videos through video grounding methods. (ii) In terms of image quality and temporary style performance, the best performance our method achieves is higher than the baseline method, indicating that it can improve the generated videos' image quality. However, these improvements are not significant, so it is worth further exploring ways to improve performance by introducing other image generation models or other temporal consistency video editing methods.

\subsection{Case Analysis}
As shown in Figure~\ref{fig:example}, we select a few video samples generated using our method and baseline method Text2Video-Zero respectively for case analysis. Our method uses VidToMe as the video editing method and selects the depth as the type of ControlNet.
We further visualize the X-T slice for each frame from the videos.

From Figure~\ref{fig:example}(a) we can see that the X-T slice of our generated video \textit{`A knight is seen riding around on a horse'} exhibits continuous subject and background changes over time, demonstrating the superiority of our method in generating stable and variable long video.
Figure~\ref{fig:example}(c) shows the X-T slice of our generated video \textit{`Iron Man is doing gymnastics in the gym'}. Despite the frequent and non-linear movements of the subject in the video, its X-T slice presents continuous gyroscopic changes, indicating our method is able to continuously generate video frames that reflect complex subject's movement patterns.
Figure~\ref{fig:example}(e) shows the X-T slice of our generated video \textit{`Iron Man lifts heavy weight over his head, in the snow'}, which includes both \textbf{customized subject} and \textbf{scenario}, demonstrating the ability of our method for multiple customization.

Compared to our method, however, Figure~\ref{fig:example}(b),(d), and (e) show the examples of corresponding generated videos and their X-T slice from our baseline method, which can be seen suffering from strong quality insufficiency, e.g., inconsistencies between frames and subject missing from partial frames. In summary, our approach of multi-sentence video grounding-based long video generation leads to more consistent video generation.

\subsection{Ablation Studies}

\subsubsection{Results with different editing method}
% We compared different LLMs, including GPT-3.5, Vicuna-7B, Longchat-7B, and some VidLLMs such as Video-Chat, Video-ChatGPT, and Video-LLaMA. For normal LLMs, we used the aforementioned combination of LLMs and visual descriptions to complete the video grounding task. For VidLLMs, we directly asked them to read the corresponding video content and answer the video grounding question. The final results are shown in Table~\ref{tab:llm}, and we can draw the following conclusion based on this.
As shown in Table~\ref{tab:total}, we compared the results of different video editing methods using different video editing approaches and different pre-trained weights of ControlNet. From the results, we can observe that:(i) The video editing method such as VidToMe, achieves overall better performance in editing long videos compared to other video editing methods. We believe this is mainly due to VidToMe not only considering the local continuity relationship between video frames but also strengthening the long-term consistency of generated videos by introducing global tokens, while other video editing methods only rely on inter-frame correction, indicating the importance of considering global temporal relationships for long video generation. (ii) applying a single type of ControlNet to edit different video segments using the same method for long video generation does not significantly improve the overall quality of the generated video. This can be attributed to the fact that editing for different types of videos is suitable for different types of ControlNets, and it is worth trying multiple ControlNets to further explore their impact on video editing in future work.

\subsubsection{Results with Video Morphing and Personalization}
As shown in Table~\ref{tab:ablation}, we compared the results of applying video morphing and personalization methods. The results are conducted under VidToMe using ControlNet in the type of depth. We can see from the result that with the introduction of video morphing and personalization approaches, our method still outperforms the baseline Text2Video-Zero model. In addition, we present an example long video generated through our method shown in Figure~\ref{fig:positive}. The generated long video maintains good consistency in the customized subject across frames.

\section{limitations and future works}
Since this is the first attempt at multi-sentence video grounding for long video generation,
the main limitation of our method lies in the model and dataset of multiple-sentence video grounding.
As shown in Figure~\ref{fig:failure1}, the model failed to search for partial text queries correctly in the video dataset (while the video segments of \textit{`A dog catches a ball'} and \textit{`A butterfly is flying on the sky'} do exist in the ActivityNet dataset), resulting in a low correlation between the searched videos and text queries and affecting further video editing.

On the other hand, the dataset lacks video clips of specific types of queries, which also causes inconvenience during the retrieval. The examples are shown in Figure~\ref{fig:failure2} since the dataset do not possess video clips related to \textit{`submarine'} or \textit{`snake'}.
% We initially conduct experiments for the Charades dataset, which mainly includes simple video scenes from indoors, many outdoor activities and complex scenes are not involved, making it difficult to retrieve diverse video clips suitable for video editing. While the video data from the TACoS dataset mainly focuses on cooking, leading to similar problems. 

These limitations can be addressed by improving the performance of video grounding models and introducing more diverse video datasets to further enhance the overall performance of our method. 
% We believe our work takes a further step towards a more 

\section{conclusion}
In this paper, we study the problem of utilizing video grounding models to conduct data augmentation for long video generation.
We bravely propose the Multi-sentence Video Grounding for Long Video Generation framework, which consists of a multi-sentence video grounding model to retrieve different video moments matching the target text queries from the video database, and a data augmentation strategy to edit video contents into videos with unified subjects through video editing method. Experiments demonstrate that our proposed framework outperforms baseline models for long video generation. Our approach seamlessly extends the development in image/video editing, video morphing, personalized generation, and video grounding to the long video generation, offering effective solutions for generating long videos at low memory cost. Utilizing video grounding methods to enhance the long video generation could be a promising future research direction.

% \begin{acks}
% \end{acks}

\bibliographystyle{ACM-Reference-Format}
\bibliography{sample-base}

\end{document}